\documentclass[TR]{ucl_document}
\usepackage{Definitions}
\usepackage{bm}
\usepackage{mlapa}
\usepackage{algorithmic}
\usepackage{algorithm}

\renewcommand{\Re}{\mathbb{R}}
\newcommand{\calA}{\mathcal{A}}
\newcommand{\calX}{\mathcal{X}}

\begin{document} 
\title{A Nonconformity Approach to Model Selection for SVMs}

\author[Hardoon et al.]{David R. Hardoon, Zakria Hussain and John Shawe-Taylor}





\persontelephone{+44 (0)20 7679 0425}
\personemail{\{D.Hardoon, Z.Hussain, jst\}@cs.ucl.ac.uk}
\personurl{//www.cs.ucl.ac.uk/staff/\{D.Hardoon, Z.Hussain, J.Shawe-Taylor\}/}

\documentnumber{}

\maketitle

\begin{abstract}
We investigate the issue of model selection and the use of the nonconformity (strangeness) measure in batch learning. Using the nonconformity measure we propose a new training algorithm that helps avoid the need for Cross-Validation or Leave-One-Out model selection strategies. We provide a new generalisation error bound using the notion of nonconformity to upper bound the loss of each test example and show that our proposed approach is comparable to standard model selection methods, but with theoretical guarantees of success and faster convergence. We demonstrate our novel model selection technique using the Support Vector Machine.
\end{abstract}

\section{Introduction}
Model Selection is the task of choosing the best model for a particular data analysis task. It generally makes a compromise between fit with the data and the complexity of the model. Furthermore, the chosen model is used in subsequent analysis of test data.  Currently the most popular techniques used by practitioners are Cross-Validation (CV) and Leave-One-Out (LOO). 


In this paper the model we concentrate on is the Support Vector Machine (SVM) \cite{Boser92}.  CV and LOO are the modus operandi despite there being a number of alternative approaches proposed in the SVM literature.
For instance, \emcite{OC_VV-99} explore model selection using the span of the support vectors and re-scaling of the feature space, whereas, \emcite{MM_KB-02}, motivated by an application in drug design, propose a fully-automated search methodology for model selection in SVMs for regression and classification.  \emcite{Gold03modelselection} give an in depth review of a number of model selection  alternatives for tuning the kernel parameters and penalty coefficient $C$ for SVMs, and although they find a model selection technique that performs well (at high computational cost), the authors conclude that ``\textit{the hunt is still on for a model selection criterion for SVM classification which is both simple and gives consistent generalisation performance}".  More recent attempts at model selection have been given by
\emcite{Hastie04} who derive an algorithm that fits the entire path of SVM solutions for every value of the cost parameter, while \emcite{HL_SW_FQ-05} propose to use the Vapnik-Chervonenkis (VC) bound; they put forward an algorithm that employs a coarse-to-fine search strategy to obtain the best parameters in some predefined ranges for a given problem.  Furthermore, \emcite{AA_EPH_JST-06} propose a tighter PAC-Bayes bound to measure the performance of SVM classifiers which in turn can be used as a way of estimating the hyperparameters. Finally, \emcite{Bruno-06} have addressed model selection for multi-class SVMs using Particle Swarm Optimisation.


Recently, \emcite{SO_ZH_JST-IP}, following on work by \emcite{SO_JST_GWW_ZBO-IP}, show that selecting a model whose hyperplane achieves the maximum separation from a test point obtains comparable error rates to those found by selecting the SVM model through CV.  In other words, while methods such as CV involve finding one SVM model (together with its optimal parameters) that minimises the CV error, \emcite{SO_ZH_JST-IP} keep all of the models generated during the model selection stage and make predictions according to the model whose hyperplane achieves the maximum separation from a test point.  The main advantage of this approach is the computational saving when compared to CV or LOO.  However, their method is only applicable to large margin classifiers like SVMs.

We continue this line of research, but rather than using the distance of each test point from the hyperplane we explore the idea of using the \emph{nonconformity measure} \cite{VV_AG_GS-05,GS_VV-08} of a test sample to a particular label set.  The nonconformity measure is a function that evaluates how `strange' a prediction is according to the different possibilities available.  
The notion of nonconformity has been proposed in the on-line learning framework of conformal prediction \cite{GS_VV-08}, and is a way of scoring how different a new sample is from a bag\footnote{A \emph{bag} is a more general formalism of a mathematical \emph{set} that allows repeated elements.} of old samples.  The premise is that if the observed samples are well-sampled then we should have high confidence on correct prediction of new samples, given that they \emph{conform} to the observations.  

We take the nonconformity measure and apply it to the SVM algorithm during testing in order to gain a time advantage over CV and to generalise the algorithm of \emcite{SO_ZH_JST-IP}.  Hence we are not restricted to SVMs (or indeed a measure of the margin for prediction) and can apply our method to a broader class of learning algorithms.  However, due to space constraints we only address the SVM technique and leave the application to other algorithms (and other nonconformity measures not using the margin) as a future research study.  Furthermore we also derive a novel learning theory bound that uses nonconformity as a measure of complexity.  To our knowledge this is the first attempt at using this type of measure to upper bound the loss of learning algorithms. 

The paper is laid out as follows. In Section \ref{sec:nom} we present the definitions used throughout the paper. Our main algorithmic contributions are given in Section \ref{sec:con} where we present our nonconformity measure and its novel use in prediction. Section \ref{sec:bound} presents a novel generalisation error bound for our proposed algorithm. Finally, we present experiments in Section \ref{sec:exp} and conclude in Section \ref{sec:dis}.


\section{Definitions}
\label{sec:nom}
The definitions are mainly taken from \emcite{GS_VV-08}.

Let $(x_i,y_i)$ be the $i$th input-output pair from an input space $\X$ and output space $\Y$.  Let $z_i = (x_i,y_i)$ denote short hand notation for each pair taken from the joint space ${\bf Z} := \X \times \Y$.

We define a \emph{nonconformity measure} as a real valued function $A(S,z)$ that measures how different a sample $z$ is from a set of observed samples $S = \{z_1,\ldots,z_m\}$.  A nonconformity measure must be fixed \emph{a priori} before any data has been observed.

Conformal predictions work by making predictions according to the nonconformity measure outlined above.  Given a set $S = \{z_1,\ldots,z_m\}$ of training samples observed over $t=1,\ldots,m$ time steps and a new sample $x$, a conformal prediction algorithm will predict $y$ from a set containing the correct output with probability $1-\ep$.  For example, if $\ep = 0.05$ then the prediction is within the so-called \emph{prediction region} -- a set containing the correct $y$, with $95\%$ probability.  In this paper, we extend this framework to the batch learning model to make predictions using confidence estimates, where for example we are $95\%$ confident that our prediction is correct.

In the batch learning setting, rather than observing samples incrementally such as $x_1, y_1, \ldots, x_m, y_m$ we have a training set $S = \{(x_1,y_1),\ldots,(x_m,y_m)\}$ containing all the samples for training that are assumed to be distributed i.i.d. from a fixed (but unknown) distribution $\Dcal$.  Given a function (hypothesis) space $\Hcal$ the batch algorithm takes training sample $S$ and outputs a hypothesis $f : \X \mapsto \Y$ that maps samples to labels.


For the SVM notation let $\phi : \X \mapsto \mathbf{F}$ map the training samples to a higher dimensional feature space $\mathbf{F}$.  The primal SVM optimisation problem can be defined like so:
\begin{eqnarray*}
\begin{array}{ll}
\min_{w,b} & \| w \|_2^2  + C \sum_{i=1}^n\xi_i \\
\mathrm{subject \ to } & y_i \left( \left<w,\phi(x_i)\right> + b \right) \geq 1 - \xi_i \\
& i = 1,\ldots,n.
\end{array}
\end{eqnarray*}
where $b$ is the bias term, $\xi \in \Reals^n$ is the vector of slack variables and $w \in \Reals^n$ is the primal weight vector, whose 2-norm minimisation corresponds to the maximisation of the margin between the set of positive and negative samples.  The notation $\langle \cdot , \cdot \rangle$ denotes the inner product.  
The dual optimisation problem gives us the flexibility of using \emph{kernels} to solve nonlinear problems \cite{BS_AS-02,ST_NC-04}.  The dual SVM optimisation problem can be formulated like so:
\begin{eqnarray*}
\begin{array}{ll}
\max_{\al} & \sum_{i}^m\alpha_i - \frac{1}{2}\sum_{i=1}^m y_i y_j \alpha_i \alpha_j \kappa(x_i,x_j),\\
\mathrm{subject\ to} & \sum_{i=1}^m y_i\alpha_i = 0,\\
& 0 \leq \alpha_i \leq C,
\end{array}
\end{eqnarray*}
where $\kappa(\cdot,\cdot)$ is the kernel function and $\alpha \in \Reals^m$ is the dual (Lagrangian) variables.  Throughout the paper we will use the dual optimisation formulation of the SVM as we attempt to find the optimal regularisation parameter for the SVM together with the optimal kernel parameters.

\section{Nonconformity Measure}
\label{sec:con}
We now discuss the main focus of the paper.  Let $S = S_{\mathrm{trn}} \cup S_{\mathrm{val}}$ be composed of a training set $S_{\mathrm{trn}}$ and a validation set $S_{\mathrm{val}}$.  We assume without loss of generality that,
\[
S = \{ z^t_1, \ldots, z^t_m, z^v_1, \ldots, z^v_n\}
\]
where $S_{\mathrm{trn}} = \{ z^t_1, \ldots, z^t_m\}$ and $S_{\mathrm{val}} = \{ z^v_1, \ldots, z^v_n\}$.  

We start by defining our nonconformity measure $A(S_{\mathrm{val}},z)$ for a function $f$ over the validation set $S_{\mathrm{val}}$ and $j=1,\ldots,n$ as,
\begin{eqnarray}
A(S_{\mathrm{val}},z) = yf(x).
\end{eqnarray}
Note that this does not depend on the whole sample but just the test point. In itself it does not characterise how different the point is. To do this we need the so called $p${\em-value} $p_A(S_{\mathrm{val}},z)$ that computes the fraction of points in $S_{\mathrm{val}}$ with `stranger' values:
\begin{eqnarray*}\label{eq:ncmeasure}
p_A(S_{\mathrm{val}}, z) = \frac{\left| \LC 1\leq j \leq n : A(S_{\mathrm{val}},z^{v}_j) \leq A(S_{\mathrm{val}},z) \RC \right|}{n},
\end{eqnarray*}
which, in this case, measures the number of samples from the validation set that have smaller functional margin than the test point functional margin. The larger the margin obtained the more confidence we have in our prediction.  The nonconformity p-value of $z$ is between $1$ and $1/n$.  If it is small (tends to $1/n$) then sample $z$ is non-conforming and if it is large (tends to $1$) then it is conforming.

\begin{figure}[htbp]
\begin{center}
\includegraphics[scale=.37]{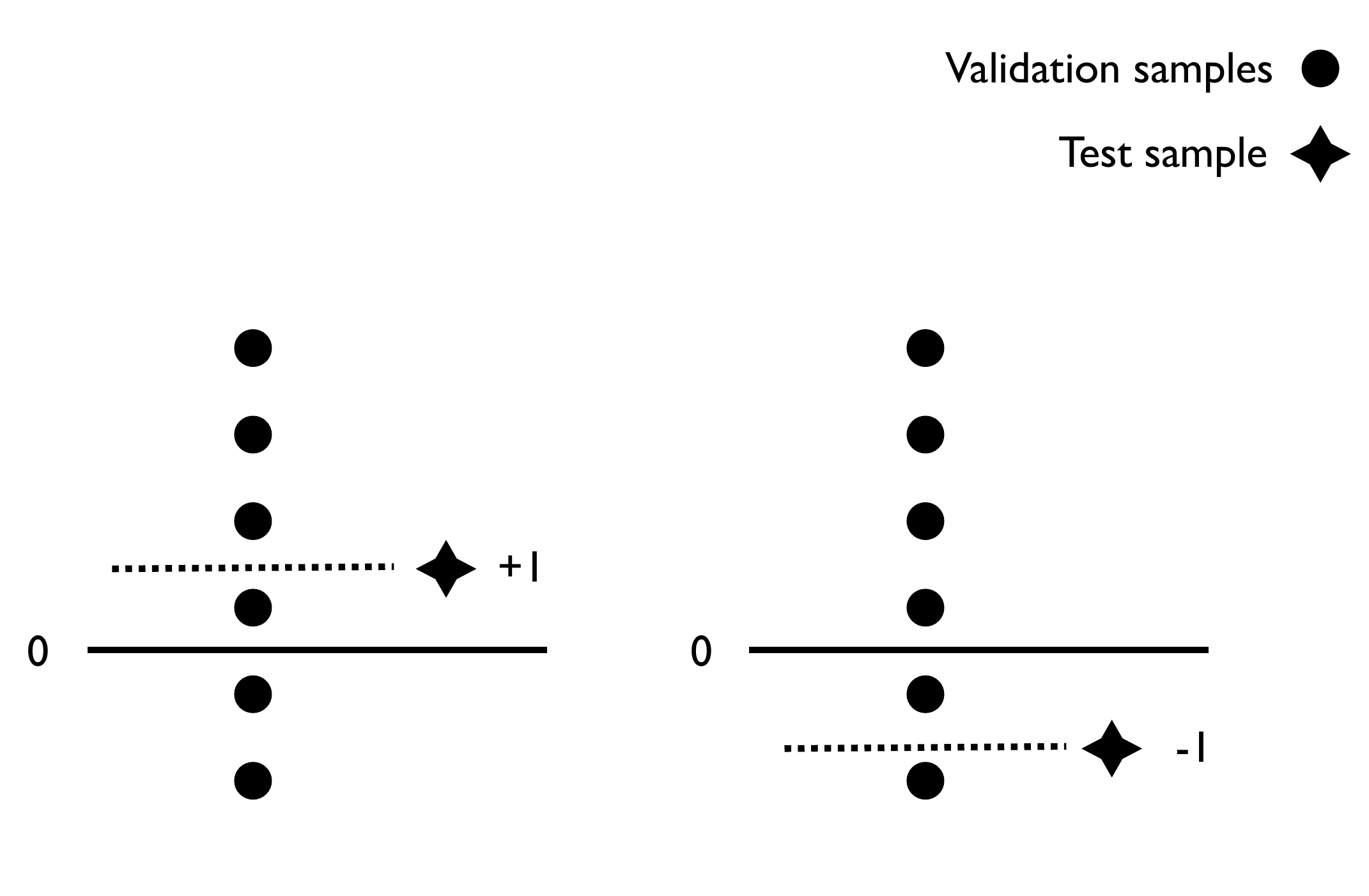}
\caption{A simple illustrative example of non-conformal prediction using a validation set of six samples (2 are misclassifications, 4 are correctly classified) on a single test sample with a positive functional value and its two label possibilities of $+1$ (left) and $-1$ (right).}
\label{fig:exa}
\end{center}
\end{figure}

In order to better illustrate this idea we show a simple pictorial example in Figure \ref{fig:exa}.  We are given six validation samples ordered around $0$ (solid line) in terms of their correct/incorrect classification \ie, the value $y^vf(x^v)$ for an $(x^v,y^v) = z^v$ pair will be correctly classified by $f$ iff $y^{v}f(x^{v}) > 0$.  In our example two are incorrectly classified (below the threshold) and four are correct.  The picture on the left also includes $yf(x)$ for a test sample $x$ when its label is considered to be positive \ie, $y=+1$.  In this case there remain $3$ validation samples below its value of $yf(x)$ giving us a nonconformity measure p-value using Equation (\ref{eq:ncmeasure}) as $p_A(S_{\mathrm{val}},(x,y=+1)) = \frac{3}{6}$.  A similar calculation can be made for the picture on the right when we consider the label $y = -1$ for test point $x$ \ie, $(x,y=-1)$.  We are able to conclude, for this sample, that assigning $x$ a label of $ y = +1$ gives a nonconformity p-value of $p_A(S_{\mathrm{val}},(x,+1)) = \frac{1}{2}$ while assigning a label of $y = -1$ gives a p-value of $p_A(S_{\mathrm{val}},(x,-1)) = \frac{1}{6}$.  Therefore, with a higher probability, our test sample $x$ is conforming to $+1$ (or equally non-conforming to $-1$) and should be predicted positive.
We state the standard result for nonconformity measures, but first define a nonconformity prediction scheme and its associated error.

\begin{definition}
For a fixed nonconformity measure $A(S,z)$, its associated $p$-value, and $\epsilon > 0$, the confidence predictor $\Gamma^\epsilon$ predicts the label set
\[
\Gamma^{\epsilon}(S, x) = \left\{y : p_A(S, (x,y)) \geq \epsilon \right\}
\]
The confidence predictor $\Gamma^\epsilon$ makes an error on sample $z = (x,y)$ if $y \not\in \Gamma^{\epsilon}(S,x)$.
\end{definition}

\begin{proposition}
\label{ncprop}
For exchangeable distributions we have that 
\[
P^{n+1}\left\{(S,z) : y \not\in \Gamma^{\epsilon}(S,x)\right\} \leq \epsilon.
\]
\end{proposition}
\newcommand{\tS}{\tilde S}
\begin{proof}
By exchangeability all permutations of a training set are equally likely. Denote with $\tS$ the set $S$ extended with the sample $z_{n+1}$ and for $\sigma$ a permutation of $n+1$ objects.  Let $\tS_\sigma$ be the sequence of samples permuted by $\sigma$.
Consider the permutations for which the corresponding prediction of the final element of the sequence is not an error. This implies that the value $A(\tS_\sigma,z_{\sigma(n+1)})$ is in the upper $1-\epsilon$ fraction of the values $A(\tS_\sigma,z_{\sigma(i)}), i=1, \ldots, n+1$. This will happen at least $1-\epsilon$ of the time under the permutations, hence upper bounding the probability of error over all possible sequences by $\epsilon$ as required.
\end{proof}

Following the theoretical motivation from \emcite{John98a} we proceed by computing all the SVM models and applying them throughout the prediction stage.  A fixed validation set, withheld from training, is used to calculate the nonconformity measures.  We start by constructing $K$ SVM models so that each decision function $f_k \in F$ is in the set $F$ of decision functions with $k=1,\ldots,K$.  The different set of SVM models can be characterised by different regularisation parameters for $C$ (or $\nu$ in $\nu$-SVM) and the width parameter $\gm$ in the Gaussian kernel case.  For instance, given $10$ $C = \{C_1,\ldots,C_{10}\}$ values and $10$ $\gm = \{\gm_1,\ldots,\gm_{10}\}$ values for a Gaussian kernel we would have a total of $|C| \times |\gm| = 100$ SVM models, where $|\cdot|$ denotes the cardinality of a set.

We now describe our new model selection algorithm for the SVM using nonconformity.  If the following
\[
\frac{\left| \LC \forall j  : y_j f_k(x_j) \leq y f_k(x) \RC \right|}{n} > \ep
\]
statement holds, then we include $y \in \Gm_k^\ep$ where $\Gm$ is the prediction region (set of labels conforming).  For classification, the set $\Gm$ can take the following values:
\[
\{ \emptyset \}, \{ -1 \} , \{ + 1 \}, \{ -1, +1 \}.
\]
Clearly finding the prediction region $\Gm = \{-1\}$ or $\Gm = \{+1\}$ is useful in the classification scenario as it gives higher confidence of the prediction being correct, while the sets $\Gm = \{ \emptyset \}$ and $\Gm = \{ -1, +1\}$ are useless as the first abstains from making a prediction whilst the second is unbiased towards a label.

Let $\ep_{crit}$ be the critical $\ep$ that creates one label in the set $\Gm_{k}^{\ep}$ for at least one of the $K$ models:
\begin{eqnarray}
\label{eqn:nonc}
\ep_{crit} = \min_{k \in K} \min_{y \in \{-1,+1\}} \frac{\left| \{ \forall j : y^v_j f_k(x^v_j) \leq y f_k(x) \} \right|}{n}.
\end{eqnarray}
Furthermore, let $k_{crit}, y_{crit}$ be arguments that realise the minimum $\ep_{crit}$, chosen randomly in the event of a tie. This now gives
the prediction of $x$ as $y = -y_{crit}$. This is because $y_{crit}$ is non-conforming (strange) and we wish to select the \emph{opposite} (conforming) label.  In the experiments section we refer to the prediction strategy outlined above and the model selection strategy given by equation (\ref{eqn:nonc}) as the nonconformity model selection strategy.  We set out the pseudo-code for this procedure in Algorithm \ref{al:alg}.

\begin{algorithm}
\begin{algorithmic}[1]
\REQUIRE Sample $S = \{ (x_i,y_i) \}^{\ell}_{i=1}$, SVM parameters $C$ and $\gm$ (for Gaussian kernel) where $K = |C| \times |\gm|$
\ENSURE Predictions of test points $x_{\ell+1},x_{\ell+2},\ldots$
\STATE Take training data $S$ and randomly split into training set $S_{\mathrm{trn}} = \{(x^t_1,y^t_1),\ldots (x^t_m,y^t_m)\}$ and validation set $S_{\mathrm{val}} = \{ (x_{1}^v,y_{1}^v),\ldots,(x_{n}^v,y_{n}^v) \}$ where $m+n = \ell$ \COMMENT{\texttt{This split is only done {\em{once}}}}.
\STATE Train $K$ SVM models on training data $S_{\mathrm{trn}}$ to find $f_1(\cdot),\ldots,f_K(\cdot)$.
\STATE {\bf Prediction Procedure}: For a test point $x$ compute: 
\[
\ep_{crit} = \min_{k \in K} \min_{y \in \{-1,+1\}} \LC \frac{\left| \{ \forall j : y^v_j f_k(x^v_j) \leq y f_k(x) \} \right|}{n} \RC,
\]
realised by $k = k_{crit}$ and $y = y_{crit}$.
\STATE Predict label $-y_{crit}$ for $x$.
\end{algorithmic}
\caption{Nonconformity model selection.}
\label{al:alg}
\end{algorithm}


Before proceeding we would like to clarify some aspects of the Algorithm.  The data is split into a training and validation set \emph{once} and therefore all $K$ models are computed on the training data -- after this procedure we only require to calculate the nonconformity measure p-value for all test points in order to make predictions.  However, in $b$-fold Cross-Validation we require to train, for each $C$ and $\gm$ parameter, a further $b$ times.  Hence CV will be at most $b$ times more computationally expensive.


\section{Nonconformity Generalisation Error Bound}
\label{sec:bound}

The problem with Proposition \ref{ncprop} is that it requires the validation set to be generated afresh for each test point, specifies just one value of $\epsilon$, and only applies to a single test function. In our application we would like to reuse the validation set for all of our test data and use an empirically determined value of $\epsilon$.  Furthermore we would like to use the computed errors for different functions in order to select one for classifying the test point.

We therefore need to have uniform convergence of empirical estimates to true values for all values of $\epsilon$ and all functions $K$. We first consider the question of uniform convergence for all values of $\epsilon$.

If we consider the cumulative distribution function $F(\gamma)$ defined by
\[
F(\gamma) = P\left((x,y) : yf(x) \leq \gamma\right),
\]
we need to bound the difference between empirical estimates of this function and its true value. This corresponds to bounding the difference between true and empirical probabilities over the sets
\[
\calA = \left\{(-\infty, a] : a \in \Re \right\}.
\]
Observe that we cannot shatter two points of the real line with this set system as the larger cannot be included in a set without the smaller. It follows that this class of functions has Vapnik-Chervonenkis (VC) dimension 1.
We can therefore apply the following standard result, see for example \emcite{DevGyoLug96}.
\begin{theorem}
\label{thm}
Let $\calX$ be a measurable space with a fixed but unknown probability distribution $P$.
Let $\calA$ be a set system over $\calX$ with VC dimension $d$ and fix $\delta > 0$. With probability at least $1 - \delta$ over the generation of an i.i.d. $m$-sample $S \subset \calX$, 
\[
\left|\frac{|S\cap \calA|}{m} - P(\calA)\right| \leq 5.66\sqrt{\frac{d\ln\left(\frac{em}{d}\right) + \ln\frac{8}{\delta}}{m}}.
\]
\end{theorem}

We now apply this result to the error estimations derived by our algorithm for the $K$ possible choices of model.

\begin{proposition}
\label{prop:bound}
Fix $\delta >0$. Suppose that the validation set $S_{val}$ of size $n$ in Algorithm~\ref{al:alg} has been chosen i.i.d. according to a fixed but unknown distribution that is also used to generate the test data. Then with probability at least $1- \delta$ over the generation of $S_{val}$, if for a test point $x$ the algorithm returns a classification $y^{v} = -y_{crit}$, using function $f_{k_{crit}}$, $1 \leq k_{crit} \leq K$, realising a minimum value of $\ep_{crit}$, then the probability of misclassification satisfies
\[
P\left((x,y) : y \not= y^{v}\right) \leq \ep_{crit} + 5.66\sqrt{\frac{\ln\left(en\right)+ \ln\frac{8K}{\delta}}{n} }.
\]
\end{proposition}

\begin{proof}
We apply Theorem \ref{thm} once for each function $f_k$, $1 \leq k \leq K$ with $\delta$ replaced by $\delta/K$. This implies that with probability $1-\delta$ the bound holds for all of the functions $f_k$, including the chosen $f_{k_{crit}}$. For this function the empirical probability of the label $y_{crit}$ being observed is $\ep_{crit}$, hence the true probability of this opposite label is bounded as required.
\end{proof}

\def\omit{
In order to achieve this type of bound we will show that the error rate distribution over the choice of the validation set $S$ is concentrated. This will ensure that with high probability selecting a single fixed validation set will give the performance that we require. We make use of McDiarmid's inequality.

\begin{theorem}
(McDiarmid \cite{McDiarmid}) Let $X_{1},\ldots ,X_{n}$ be
independent random variables taking values in a set $A$, and
assume that $f:A^{n}\rightarrow \mathbb{R}$ satisfies
\begin{eqnarray*}
\sup_{x_{1},\ldots ,x_{n},x_{i}^{\prime }\in A}|f(x_{1},\ldots
,x_{n})&-&\\
&& \hspace*{-1.5cm} f(x_{1},\ldots ,x_{i-1},x_{i}^{\prime },x_{i+1},\ldots
,x_{n})| \\
&\leq&
c_{i},%
\text{ \ }1\leq i\leq n.
\end{eqnarray*}%
Then for all $\epsilon >0$,
\begin{equation}
P\{f(X_{1},\ldots ,X_{n})-{\mathbb{E}}f \geq
\epsilon \}\leq \exp \left( {\frac{-2\epsilon
^{2}}{\sum_{i=1}^{n}c_{i}^{2}}}\right)
\end{equation}
\end{theorem}
We apply McDiarmid's inequality where $f(z_1, \ldots, z_{n+1}) = p_A(S, z_{n+1})$.
In order to apply McDiarmid's inequality we need to show that replacing one of the examples in the validation set will not affect the error rate
}

\begin{remark}
The bound in Proposition \ref{prop:bound} is applied \emph{using} each test sample which in turn gives a different bound value for each test point (\eg, see \emcite{John98a}).  Therefore, we are unable to compare this bound with existing training set CV bounds \cite{MK_DR-99,TZ-01} as they are traditional a \emph{priori} bounds computed over the training data, and which give a uniform value for all test points (\ie, training set bounds \cite{Langford05}).
\end{remark}

\section{Experiments}
\label{sec:exp}

In the following experiments we compare SVM model selection using traditional CV to our proposed nonconformity strategy as well as to the model selection using the maximum margin \cite{SO_ZH_JST-IP} from a test sample.

We make use of the Votes, Glass, Haberman, Bupa, Credit, Pima, BreastW and Ionosphere data sets acquired from the UCI machine learning repository.\footnote{http://archive.ics.uci.edu/ml/} The data sets were pre-processed such that samples containing unknown values and contradictory labels were removed. Table \ref{tab:data} lists the various attributes of each data set. The LibSVM package 2.85
\cite{CC01a} and the Gaussian kernel were used throughout the experiments.
\begin{table*}[htd]
\caption{Description of data sets: Each row contains the name of the data set, the number of samples and features (i.e. attributes) as well as the total number of positive and negative samples. }
\begin{center}
\begin{tabular}{|c||c|c|c|c|}
\hline
Data set & \# Samples & \# Features & \# Positive Samples & \# Negative Samples\\
\hline
Votes & 52 & 16 & 18 & 34\\
Glass & 163 & 9 & 87 & 76\\
Haberman & 294 & 3 & 219 & 75\\
Bupa & 345 & 6 & 145 & 200\\
Credit & 653 & 15 & 296 & 357\\
Pima & 768 & 8 & 269 & 499\\
BreastW & 683 & 9 & 239 & 444\\
Ionosphere& 351 & 34 & 225 & 126\\
\hline
\end{tabular}
\end{center}
\label{tab:data}
\end{table*}%
Model selection was carried out for the values listed in Table \ref{tab:values}.
\begin{table*}[htd]
\caption{Model selection values for $\gamma$ and $C$ for both cross-validation and nonconformity measure.}
\begin{eqnarray*}
\gamma &=& \{2^{-15},2^{-13},2^{-11},2^{-9},2^{-7},2^{-5} ,2^{-3},2^{-1},2^{1},2^{3}\}\\
C &=  &  \{2^{-5},2^{-3},2^{-1},2^{1},2^{3}, 2^{5}, 2^{7},2^{9}, 2^{11},2^{13},2^{15}\}
\end{eqnarray*}
\label{tab:values}
\end{table*}%

In the experiments we apply a 10-fold CV routine where the data is split into 10 separate folds, with 1 used for testing and the remaining 9 split into a training and validation set. 
We then use the following procedures for each of the two model selection strategies:
\begin{itemize}
\item \emph{Nonconformity}: split the samples into a training and validation set of size $\min(\frac{1}{5}\ell,50)$ where $\ell$ is the number of samples.\footnote{The size of the validation set was varied without much difference in generalisation error.}  Using the training data we learn all models using $C$ and $\gm$ from Table~\ref{tab:values}.  
\item \emph{Cross-Validation}: carry out a 10-fold CV \emph{only} on the training data used in the Nonconformity procedure to find the optimal $C$ and $\gm$ from Table~\ref{tab:values}.
\end{itemize} 

The validation set is excluded from training in both methods, but used for prediction in the nonconformity method. Hence, the samples used for training and testing were identical for both CV and the nonconformity model selection strategy.  We feel that this was a fair comparison as both methods were given the same data samples from which to train the models. 

Table \ref{tab:res} presents the results where we report the average error and standard deviation for Cross-Validation and the nonconformity strategy.  We are immediately able to observe that carrying out model selection using the nonconformity measure is, on average, a factor of $7.3$ times faster than using CV. The results show that (excluding the Haberman data set) nonconformity seems to perform similarly to CV in terms of generalisation error.  However, lower values for the standard deviation on Votes, Glass, Bupa and Credit suggest that on these data sets nonconformity gives more consistent results than CV.  Furthermore, when excluding the Haberman data set, the overall error for the model selection using nonconformity is $0.1730\pm0.0659$ and CV is $0.1686\pm0.0886$, constituting a difference of only $0.0044$ (less than half a percent) in favour of CV and a standard deviation of $0.0227$ in favour of the nonconformity approach.  We hypothesise that the inferior results for Haberman are due to the very small numbers of features (only $3$).  

We also compare the nonconformity strategy to the SVM $L_\infty$ maximum margin approach \cite{SO_ZH_JST-IP}. The SVM $L_\infty$ selects the model with the maximum margin from the test sample in order to make predictions. Once again, the training and testing sets were identical for both methods.  Observe that despite the $L_\infty$ being approximately $7$s faster (on average) than our proposed method, we obtain an improvement of $0.0251\pm0.0108$.  Hence, bringing us closer to the CV error rate (nonconformity is overall only $1.17\%$ worse than CV when including the Haberman dataset and 0.44\% worse when excluding).  In fact we obtain lower error rates, than SVM $L_\infty$, on all datasets except for Credit (but with a smaller standard deviation).

\begin{sidewaystable}
\caption{Model selection results: Average error and standard deviation as well as the run-time (in seconds) for model selection using nonconformity measure, 10-fold Cross Validation and the SVM-$L_\infty$ margin distance .}
\begin{tabular}{|c||c|c||c|c||c|c|}
\hline
Data set & Nonconformity & Run-Time & Cross-Validation & Run Time &SVM-$L_\infty$ & Run-Time\\
\hline
Votes & $0.0700\pm0.1201$ & $0.72$s & $0.0833\pm0.2115$ & $5.74$s& $0.0933\pm0.0991$ & $0.43$s\\
Glass & $0.2167\pm0.0932$ & $5.25$s & $0.2085\pm0.1291$ & $32.58$s& $0.2328\pm0.1263$ & $3.08$s\\
Haberman & $0.3133\pm0.0680$ & $58.23$s& $0.2518\pm0.0397$ & $455.77$s& $0.3300\pm0.0524$ & $49.23$s\\
Bupa & $0.2753\pm0.0620$ & $44.85$s & $0.2840\pm0.0604$ & $329.96$s& $0.3192\pm0.1085$ & $38.81$s\\
Credit & $0.2990\pm0.0468$ & $86.85$s & $0.2745\pm0.1111$ & $592.42$s & $0.2914\pm0.0850$ & $72.31$s\\
Pima & $0.2562\pm0.0554$ & $169.05$s& $0.2473\pm0.0361$ & $1305.29$s& $0.3019\pm0.0516$ & $155.65$s\\
BreastW & $0.0378\pm0.0350$& $24.80$s & $0.0335\pm0.0282$ & $150.36$s& $0.0408\pm0.0367$ & $18.29$s\\
Ionosphere & $0.0562\pm0.0493$& $17.63$s& $0.0479\pm0.0440$ & $103.45$s& $0.1158\pm0.0565$ & $11.85$s\\
\hline
Overall & $0.1905\pm0.0662$ & $50.92$s & $0.1788\pm0.0825$ & $371.94$s& $0.2156\pm0.0770$ & $43.71$s \\
\hline \hline
Overall ex. Haberman & $0.1730\pm0.0659$ & $49.87$s & $0.1686\pm0.0886$& $359.97$s & $0.1993\pm0.0805$& $42.91$s\\
\hline
\end{tabular}
\label{tab:res}
\end{sidewaystable}

Since we do not have a single number for the bound on generalisation (as traditional bounds) but rather individual values for each test sample, it is not possible to simply compare the bound with the test error. In order to show how the bound performs we plot the generalisation error as a function of the bound value. 

For each value of the bound we take the average error of all test points with predicted error less than or equal to that value. In other words, we create a set\footnote{Hence, no repetition of identical bound values are allowed.} $B$ containing the various bound values computed on the test samples. Subsequently, for each element in the set \ie, $\forall i, b_i\in B$ we compute the average error value for the test samples that have a bound value that is smaller or equal to $b_i$.

\begin{figure}[htbp]
\begin{center}
\includegraphics[scale=.37]{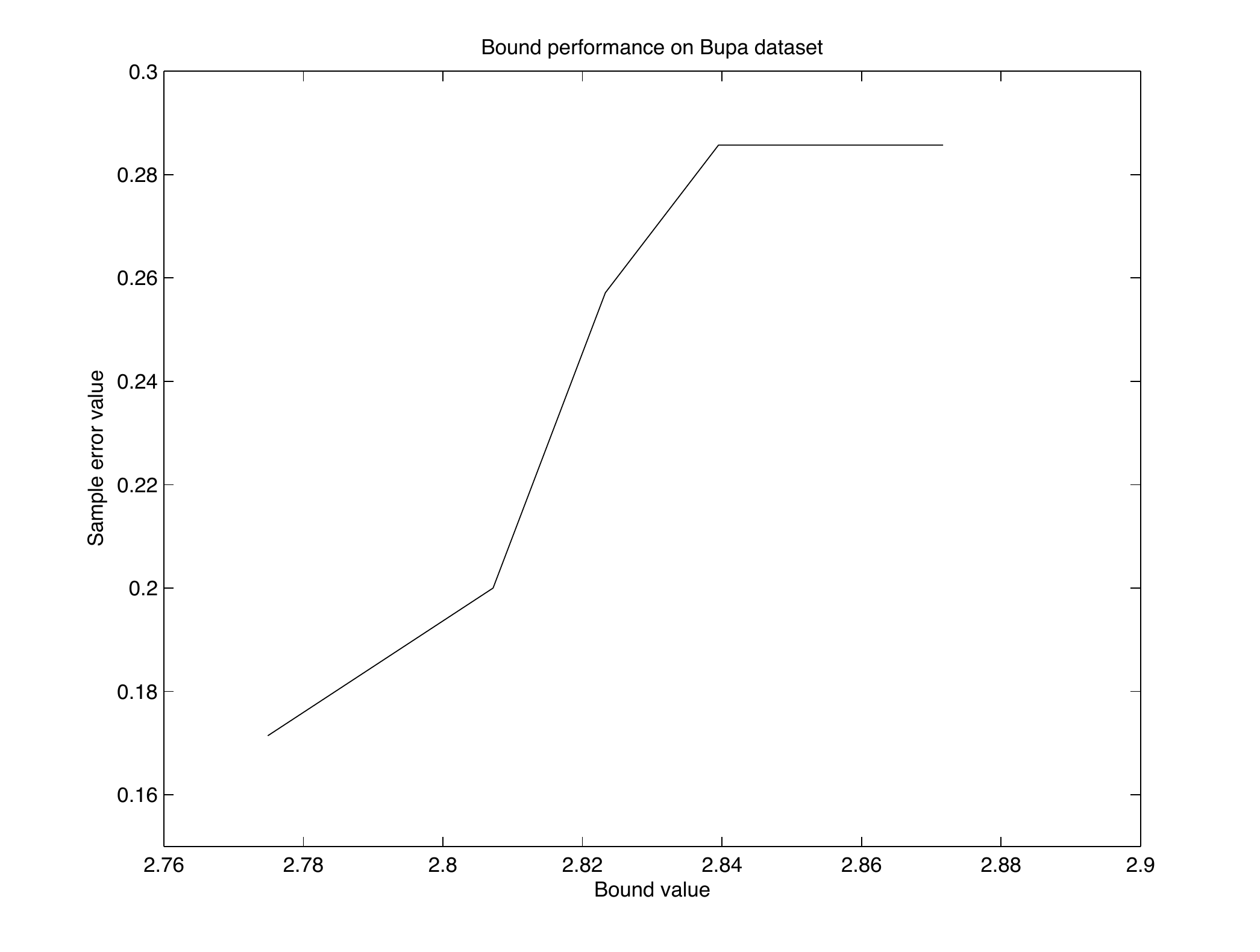}
\caption{The generalisation error as a function of the bound value for a single train-test split of the Bupa data set. The final value of the function is the overall generalisation error.}
\label{fig:bnd}
\end{center}
\end{figure}

Figure \ref{fig:bnd} shows a plot of this error rate as a function of the bound value. The final value of the function is the overall generalisation error, while the lower error rates earlier in the curve are those attainable by filtering at different bound values. As expected the error increases monotonically as a function of the bound value. Clearly there is considerable weakness in the bound, but this is partly a result of our using a quite conservative VC bound -- our main aim here is to show that the predictions are correlated with the actual error rates.

We believe these results to be encouraging as our theoretically motivated model selection technique is faster and achieves similar error rates to Cross-Validation, which is generally considered to be the gold standard.  We also find that the nonconformity strategy is slightly slower than the maximum margin approach but performs better in terms of generalisation error.

\section{Discussion}
\label{sec:dis}
We have presented a novel approach for model-selection and test sample prediction using a nonconformity (strangeness) measure. Furthermore we have given a novel generalisation error bound on the loss of the learning method. The proposed model selection approach is both simple and gives consistent generalisation performance \cite{Gold03modelselection}.

We find these results encouraging as it constitutes a much needed shift from costly model selection based approaches to a faster method that is competitive in terms of generalisation error.  Furthermore, in relation to the work of \emcite{SO_ZH_JST-IP} we have presented a method that is 1) not restricted to SVMs and 2) can use measures other than the margin to make predictions.
Therefore the nonconformity measure approach gives us a general way of choosing to make predictions, allowing us the flexibility to apply it to algorithms that are not based on large margins.  In future work we aim to investigate the applicability of our proposed model selection technique to other learning methods.  Another future research direction is to apply different nonconformity measures to the SVM algorithm presented in this paper such as, for example, a nearest neighbour nonconformity measure \cite{GS_VV-08}.

\subsubsection*{Acknowledgements}
The authors would like to acknowledge financial support from the EPSRC project Le Strum\footnote{http://www.lestrum.org}, EP-D063612-1 and from the EU project PinView\footnote{http://www.pineview.eu}, FP7-216529.


\begin{thebibliography}{}

\bibitem[Ambroladze et~al.\/, 2006][Ambroladze et~al.\/][2006]{AA_EPH_JST-06}
Ambroladze, A., Parrado-Hern{\'{a}}ndez, E., \& Shawe-Taylor, J. (2006).
\newblock Tighter {PAC-B}ayes bounds.
\newblock {\em Proceedings of Advances in Neural Information Processing
  Systems}.

\bibitem[Boser et~al.\/, 1992][Boser et~al.\/][1992]{Boser92}
Boser, B.~E., Guyon, I.~M., \& Vapnik, V.~N. (1992).
\newblock A training algorithm for optimal margin classifiers.
\newblock {\em Proceedings of the Fifth Annual Workshop on Computational
  Learning Theory} (pp.\/ 144--152).
\newblock Pittsburgh ACM.

\bibitem[Chang \& Lin, 2001][Chang and Lin][2001]{CC01a}
Chang, C.-C., \& Lin, C.-J. (2001).
\newblock {\em {LIBSVM}: a library for support vector machines}. Software
  available at {http://www.csie.ntu.edu.tw/$\sim$cjlin/libsvm}.

\bibitem[Chapelle \& Vapnik, 1999][Chapelle and Vapnik][1999]{OC_VV-99}
Chapelle, O., \& Vapnik, V.~N. (1999).
\newblock Model selection for support vector machines.
\newblock {\em Proceedings of Advances in Neural Information Processing Systems
  12} (pp.\/ 230--237).

\bibitem[de~Souza et~al.\/, 2006][de~Souza et~al.\/][2006]{Bruno-06}
de~Souza, B.~F., de~Carvalho, A. C. P. L.~F., Calvo, R., \& Ishii, R.~P.
  (2006).
\newblock Multiclass {SVM} model selection using particle swarm optimization.
\newblock {\em Proceedings of the Sixth International Conference on Hybrid
  Intelligent Systems}.

\bibitem[Devroye et~al.\/, 1996][Devroye et~al.\/][1996]{DevGyoLug96}
Devroye, L., Gy\"orfi, L., \& Lugosi, G. (1996).
\newblock {\em A probabilistic theory of pattern recognition}.
\newblock No.~31 in Applications of Mathematics. New York: Springer.

\bibitem[Gold \& Sollich, 2003][Gold and Sollich][2003]{Gold03modelselection}
Gold, C., \& Sollich, P. (2003).
\newblock Model selection for support vector machine classification.
\newblock {\em Neurocomputing}, {\em 55}, 221--249.

\bibitem[Hastie et~al.\/, 2004][Hastie et~al.\/][2004]{Hastie04}
Hastie, T., Rosset, S., Tibshirani, R., \& Zhu, J. (2004).
\newblock The entire regularization path for the support vector machine.
\newblock {\em Journal of Machine Learning Research}, {\em 5}, 1391--1415.

\bibitem[Kearns \& Ron, 1999][Kearns and Ron][1999]{MK_DR-99}
Kearns, M., \& Ron, D. (1999).
\newblock Algorithmic stability and sanity-check bounds for leave-one-out
  cross-validation.
\newblock {\em Neural Computation}, {\em 11(6)}, 1427--1453.

\bibitem[Langford, 2005][Langford][2005]{Langford05}
Langford, J. (2005).
\newblock Tutorial on practical prediction theory for classification.
\newblock {\em Journal of Machine Learning Research}, {\em 6}, 273--306.

\bibitem[Li et~al.\/, 2005][Li et~al.\/][2005]{HL_SW_FQ-05}
Li, H., Wang, S., \& Qi, F. (2005).
\newblock {SVM} model selection with the {VC} bound.
\newblock {\em Computational and Information Science}, {\em 3314}, 1067--1071.

\bibitem[Momma \& Bennett, 2002][Momma and Bennett][2002]{MM_KB-02}
Momma, M., \& Bennett, K.~P. (2002).
\newblock Pattern search method for model selection of support vector
  regression.
\newblock {\em Proceedings of the Second SIAM International Conference on Data
  Mining}.

\bibitem[{\"{O}}z{\"{o}}{\u{g}}{\"{u}}r et~al.\/,
  2008][{\"{O}}z{\"{o}}{\u{g}}{\"{u}}r et~al.\/][2008]{SO_JST_GWW_ZBO-IP}
{\"{O}}z{\"{o}}{\u{g}}{\"{u}}r, S., Shawe-Taylor, J., Weber, G.~W., \&
  {\"{O}}gel, Z.~B. (2008).
\newblock Pattern analysis for the prediction of fungal pro-peptide cleavage
  sites.
\newblock {\em Discrete Applied Mathematics, Special Issue on Networks in
  Computational Biology}, {\em doi:10.1016/j.dam.2008.06.043}.

\bibitem[{\"{O}}z{\"{o}}{\u{g}}{\"{u}}r-Aky{\"{u}}z et~al.\/, In
  Press][{\"{O}}z{\"{o}}{\u{g}}{\"{u}}r-Aky{\"{u}}z et~al.\/][In
  Press]{SO_ZH_JST-IP}
{\"{O}}z{\"{o}}{\u{g}}{\"{u}}r-Aky{\"{u}}z, S., Hussain, Z., \& Shawe-Taylor,
  J. (In Press).
\newblock Prediction with the {SVM} using test point margins.
\newblock {\em Annals of Information Systems, Special Issue on Optimization
  methods in Machine Learning}.

\bibitem[Sch\"{o}lkopf \& Smola, 2002][Sch\"{o}lkopf and Smola][2002]{BS_AS-02}
Sch\"{o}lkopf, B., \& Smola, A. (2002).
\newblock {\em Learning with kernels}.
\newblock Cambridge, MA: MIT Press.

\bibitem[Shafer \& Vovk, 2008][Shafer and Vovk][2008]{GS_VV-08}
Shafer, G., \& Vovk, V. (2008).
\newblock A tutorial on conformal prediction.
\newblock {\em Journal of Machine Learning Research}, {\em 9}, 371--421.

\bibitem[Shawe-Taylor, 1998][Shawe-Taylor][1998]{John98a}
Shawe-Taylor, J. (1998).
\newblock Classification accuracy based on observed margin.
\newblock {\em Algorithmica}, {\em 22}, 157--172.

\bibitem[Shawe-Taylor \& Cristianini, 2004][Shawe-Taylor and
  Cristianini][2004]{ST_NC-04}
Shawe-Taylor, J., \& Cristianini, N. (2004).
\newblock {\em Kernel methods for pattern analysis}.
\newblock Cambridge, U.K.: Cambridge University Press.

\bibitem[Vovk et~al.\/, 2005][Vovk et~al.\/][2005]{VV_AG_GS-05}
Vovk, V., Gammerman, A., \& Shafer, G. (2005).
\newblock {\em Algorithmic learning in a random world}.
\newblock New York: Springer.

\bibitem[Zhang, 2001][Zhang][2001]{TZ-01}
Zhang, T. (2001).
\newblock A leave-one-out cross validation bound for kernel methods with
  application in learning.
\newblock {\em Lecture Notes in Computer Science: 14th Annual Conference on
  Computation Learning Theory}, {\em 2111}, 427--443.

\end{thebibliography}
\bibliographystyle{mlapa}


\end{document}